\documentclass[letterpaper, 10 pt, journal, twoside]{IEEEtran}
\usepackage{cite}

\ifCLASSINFOpdf
\else
\fi
\newcommand{\model}{DreamUp3D}
\usepackage{graphics} 
\usepackage{amsmath} 
\usepackage{bbm}
\usepackage{amsfonts}
\usepackage{graphicx}
\usepackage{subcaption}
\usepackage{cleveref}
\usepackage{graphicx}
\usepackage[export]{adjustbox}
\usepackage{xcolor}
\renewcommand{\baselinestretch}{0.9}
\usepackage{dsfont}
\usepackage{url}

\usepackage{duckuments} 

\usepackage{booktabs}

\begin{document}
%
\title{\model: Object-Centric Generative Models for Single-View 3D Scene Understanding and Real-to-Sim Transfer}
%
%
%

\author{Yizhe Wu$^{1}$, Haitz Sáez de Ocáriz Borde$^{1}$, Jack Collins$^{1}$, Oiwi Parker Jones$^{1}$,Ingmar Posner$^{1}$%
\thanks{Manuscript received: September, 13, 2023; Revised November, 20, 2023; Accepted January, 18, 2024.}
\thanks{This paper was recommended for publication by Editor Cesar Cadena Lerma upon evaluation of the Associate Editor and Reviewers' comments.
This work was supported by the EPSRC Programme Grant (EP/V000748/1), an Amazon Research Award and the China Scholarship Council.} 
\thanks{$^{1}$All Authors are with Applied AI Lab, Oxford Robotics Institute
        {\tt\footnotesize ywu0430@outlook.com}}%
\thanks{Digital Object Identifier (DOI): see top of this page.}
}
%
%

\markboth{IEEE Robotics and Automation Letters. Preprint Version. Accepted. January, 2024}
{Wu \MakeLowercase{\textit{et al.}}: DreamUp3D} 

%



\maketitle

\begin{abstract}
3D scene understanding for robotic applications exhibits a unique set of requirements including real-time inference, object-centric latent representation learning, accurate 6D pose estimation and 3D reconstruction of objects. Current methods for scene understanding typically rely on a combination of trained models paired with either an explicit or learnt volumetric representation, all of which have their own drawbacks and limitations. We introduce \emph{\model}, a novel Object-Centric Generative Model (OCGM) designed explicitly to perform inference on a 3D scene informed only by a single RGB-D image. \emph{\model} is a self-supervised model, trained end-to-end, and is capable of segmenting objects, providing 3D object reconstructions, generating object-centric latent representations and accurate per-object 6D pose estimates. We compare \emph{\model} to baselines including NeRFs, pre-trained CLIP-features, ObSurf, and ObPose, in a range of tasks including 3D scene reconstruction, object-matching and object pose estimation. Our experiments show that our model outperforms all baselines by a significant margin in real-world scenarios displaying its applicability for 3D scene understanding tasks while meeting the strict demands exhibited in robotics applications.

\end{abstract}


%
\IEEEpeerreviewmaketitle

%
%
%
%
\begin{figure*}[hbtp!]
    \centering
    \includegraphics[width=\linewidth]{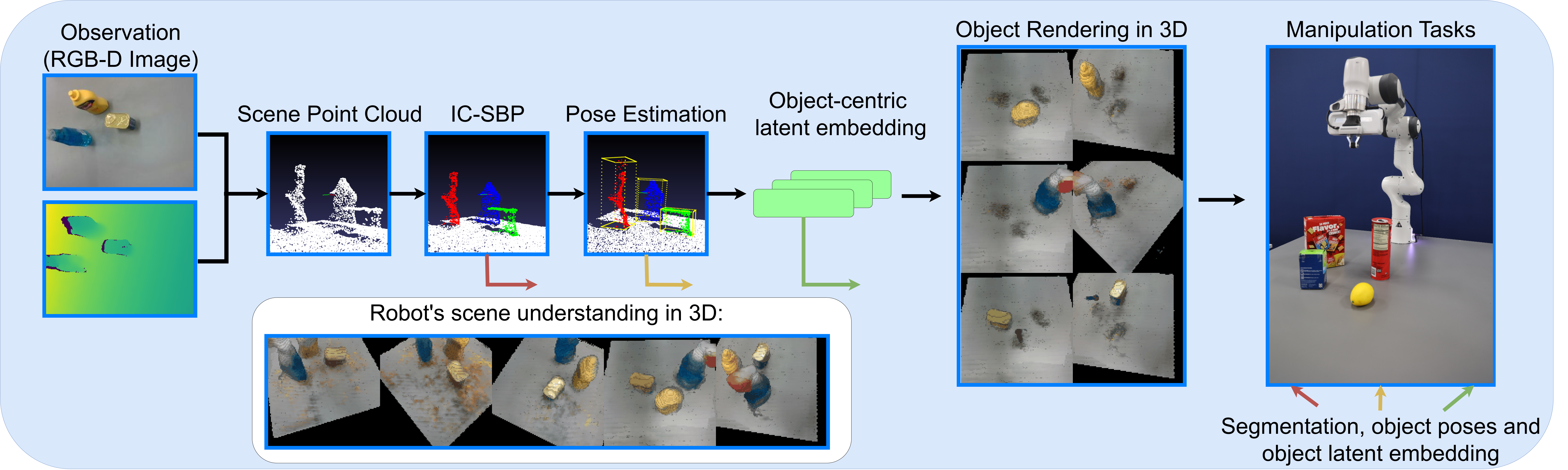}
    \caption{\model~interprets a single view RGB-D image in a 3D object-centric manner. It uses the IC-SBP algorithm to cluster the input point cloud into object masks. Then, it infers the object pose and the object-centric latent representation for each detected object. Each object is reconstructed in 3D and then together with the other objects and background component merged to form the entire 3D scene reconstruction. Predictions can then be leveraged for downstream manipulation tasks.}
    \label{fig:teaser}
    \vspace{-0.1cm}
\end{figure*}

\section{Introduction}
\IEEEPARstart{R}{obots} deployed in the real world face unique challenges as agents operating in unstructured 3D environments with only partial observability. For this reason, 3D scene understanding from limited observations is of paramount importance in facilitating tasks such as real-to-sim transfer and object manipulation. In online applications, observations need to be processed frequently through the 3D perception system to react to changes in the scene. Additionally, task-level planning requires a representation of the scene that is typically object-centric. To this end, the \textit{desiderata} for robots operating in these settings are real-time operation, 3D reconstruction based on single-viewpoint observations, object-centric latent representations, and accurate per-object 6D pose. 

Currently, NeRFs~\cite{mildenhall2020nerf, Gao2022NeRFNR} are employed as an implicit representation for 3D scene perception and understanding, and they have been widely explored in robotics. Recent studies have tested their capabilities in robotics for grasping~\cite{kerrevo}, localisation~\cite{liu2023nerf} and tactile sensing~\cite{zhong23}. However, NeRFs face significant limitations as they are formulated to represent the environment as a unified entity~\cite{kerrevo,abou2022implicit}. In particular, for each new scene a new NeRF needs to be trained adding significant overhead in time and compute. Extensions of the original NeRF formulation have reduced the impact of these limitations by leveraging multiresolution hash encoding to reduce the training time of NeRFs~\cite{muller2022instant} and by extending NeRFs to reason about individual objects in a scene \cite{yang2021objectnerf}. However, training NeRFs requires collecting images from multiple views with maximally varied viewing angles. In addition to this, identifying objects within a scene also relies on known object masks. Therefore, the NeRF formulation fails to stand up to the \textit{desiderata} for real time robot operations in unstructured environments. 

In comparison to NeRFs, OCGMs are inherently object-centric and operate in real time at inference. OCGMs leverage different attention mechanisms~\cite{Locatello2020ObjectCentricLW,Engelcke2021GENESISV2IU,lin2020space} that promote disentanglement for scene decomposition. The individual components are subsequently encoded into object-centric latent representations and decoded to reconstruct the scene. OCGMs can also be combined with NeRFs to model 3D scenes~\cite{Yu2021UnsupervisedDO}. Recent advancements have enhanced the learning process by incorporating depth supervision~\cite{Stelzner2021Decomposing3S}, enabling the training of NeRFs using RGB-D inputs without the need for explicit ray marching. However, this approach encodes both object appearance and spatial information into a single latent representation, and thus object poses cannot be inferred independently. On the contrary, ObPose~\cite{Wu2022ObPoseLC} proposes finding the minimum volume bounding box containing the object and uses the pose of the bounding box as the object pose. This enables pose estimation without labelled supervision. However, this shape-based pose estimation can be inaccurate under occlusions. Moreover, these previously proposed 3D OCGMs have only been evaluated in simulated 3D scenes, leaving the investigation of their applicability to real-world scenes to future work.

In this paper, we propose~\model, an OCGM that integrates generative radiance fields (GRAFs)~\cite{mildenhall2020nerf,schwarz2020graf,niemeyer2021giraffe} to enhance 3D scene comprehension by overcoming the limitations of past models. \model~demonstrates transfer across real-world environments and overcomes occlusions via a shape completion module which is trained by a shape distillation mechanism that reuses the GRAF predictions as a training signal. This significantly reduces computational requirements by minimising the need for repetitive evaluations of the NeRF model to retrieve object shapes, which traditionally involves thousands of evaluations using ray marching. Experimentally, we evaluate \model~against the \textit{desiderata} for 3D scene understanding in robotics tasks. Concretely, we evaluate \model~in scene reconstruction, object-centric representation learning, and pose estimation.

\section{Related Work}

Recently, NeRFs have shown their potential as a compact 3D representation of the environment and are currently being applied to many robotics tasks, such as reinforcement learning~\cite{zhou2023nerf}, SLAM~\cite{sucar2021imap}, and for grasping transparent objects~\cite{kerrevo,ichnowski2021dex}. Among NeRF implementations, instant-NGP~\cite{muller2022instant} is commonly chosen due to its fast reconstruction speed. However, these approaches often require retraining the NeRF model before each grasp to update the environment states. GraspNeRF~\cite{dai2022graspnerf} addresses this constraint by proposing a generalisable NeRF that is free from per-scene optimisation. Nevertheless, GraspNeRF is not object-centric and thus cannot interpret the scene at the object level. In contrast,~\cite{blukis2022neural} proposes to decode grasping from a learned object-centric latent representation, but the method lacks the ability to decompose a scene and requires a priori knowledge of object poses and the number of objects in the scene. Considering the desiderata for scene understanding for robots operating in the real-world, NeRFs, despite their ability to provide high-fidelity 3D reconstructions given multi-view observations, fundamentally face limitations in terms of object-level inference and real-time operation. 

OCGMs are another type of scene understanding model that offer an unsupervised method for learning latent representations at the object level, enabling reasoning at a higher level of abstraction instead of at the pixel or point cloud level. In particular, they rely on inductive biases to promote the decomposition of a scene into its individual components. This enables the models to better understand the underlying structure of a scene and capture the relationships between its constituent objects. Early works have conducted unsupervised scene inference and generation in 2D (MONet~\cite{Burgess2019MONetUS}, Slot Attention~\cite{Locatello2020ObjectCentricLW}, GENESIS~\cite{Engelcke2019GENESISGS}, GENESIS-V2~\cite{Engelcke2021GENESISV2IU}), and for robotics applications using APEX~\cite{wu2021apex,yamada2023efficient}. In both~\cite{wu2021apex,yamada2023efficient} the 2D OCGM, APEX, is utilised for object matching using the learned object-centric latent representation in an object rearrangement task in simulation and a peg-in-hole task in the real world respectively. Nevertheless, the 2D reconstruction and 2D bounding boxes predicted by such OCGM are of limited use in a 3D world. Recent research~\cite{Yu2021UnsupervisedDO} has thus focused on combining the 3D representation power of NeRFs with the versatility of OCGMs. ObSuRF~\cite{Stelzner2021Decomposing3S} proposes to further accelerate training by leveraging the depth channel of RGB-D images to guide the sampling of the NeRF queries. However, ObSuRF encodes the object appearance and location jointly into a single latent representation, and thus cannot infer per-object pose. ObPose~\cite{Wu2022ObPoseLC} addresses this problem by introducing a minimum volume principle for shape-based 6D pose estimation without using human labels. ObPose is thus the first 3D OCGM that fulfils the desiderata for scene understanding by providing real-time inference, object segmentation and 3D reconstruction, object-centric latent representation learning and unsupervised 6D pose estimation, as \model~does. ObPose also demonstrates superior performance in terms of segmentation accuracy compared to ObSuRF on the YCB~\cite{calli2015ycb}, MultiShapeNet~\cite{Stelzner2021Decomposing3S}, and CLEVR datasets~\cite{johnson2017clevr}. We therefore choose ObPose as the main baseline for the unsupervised pose estimation task.

\section{\model}
\label{method}
\begin{figure*}[htbp!]
\centering
\includegraphics[width=1.0\linewidth]{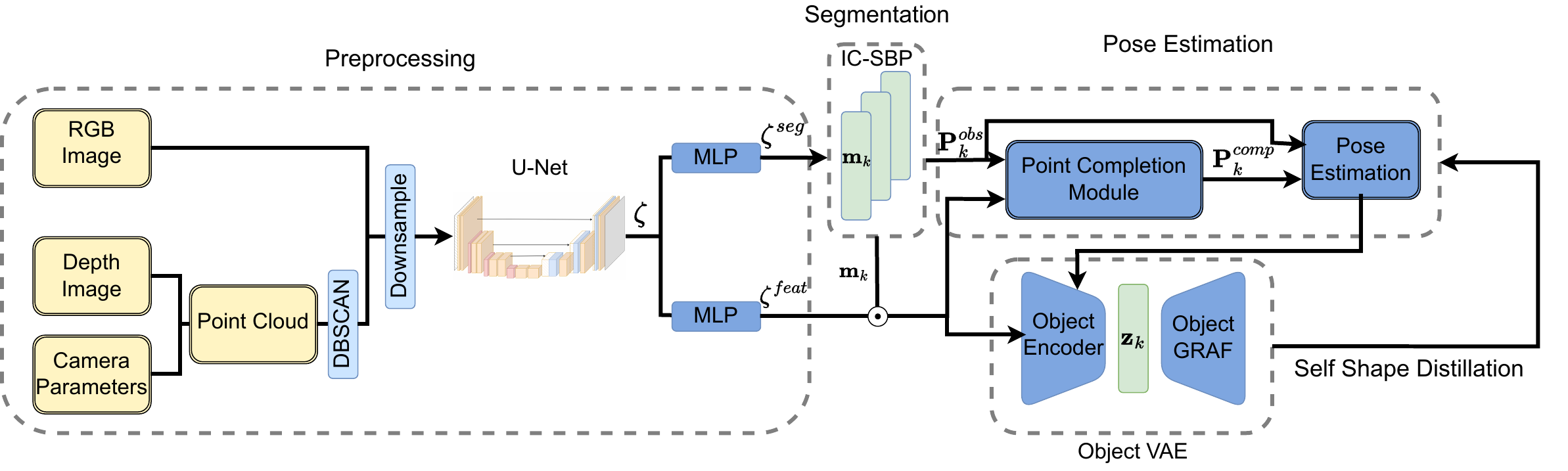} 
\caption{Architectural diagram of \model. The model is composed of several distinct modules for data preprocessing, scene segmentation, pose estimation and object encoding. See Section~\ref{method} for details.}
\label{fig:model_architecture}
\vspace{-0.15cm}
\end{figure*}

The following subsections divide the model into modules with distinct functions, starting with data preprocessing, scene segmentation for extracting object masks, pose estimation with shape completion for improved estimates, GRAF representation of objects, and concluding with an overview of model training. The overall model pipeline is depicted in Fig.~\ref{fig:teaser}. Additionally, an architectural diagram of our model can be found in Fig.~\ref{fig:model_architecture}.


\subsection{Data Pre-processing}
Given a single RGB-D image input, $\mathbf{x} \in \mathbb{R}^{\mathbf{H}\times \mathbf{W}\times 4}$, we utilize the depth information to convert the input into a point cloud. Depth images are known to be noisy, especially at object edges or when there is reflectance from metal surfaces. Therefore, we employ a clustering algorithm, DBSCAN~\cite{ester1996density}, to filter out noisy point observations from the raw input. The point cloud is then downsampled to a fixed number of points, $N$, and used, along with the RGB color, as input to a U-Net-like backbone module \cite{ronneberger2015u}, which consists of several KPConv layers \cite{thomas2019kpconv}. The KPConv layer extends the standard 2D convolutional layer to preserve the translation invariance of point cloud inputs. The output encoding produced by the U-Net, denoted as $\zeta \in \mathbb{R}^{N\times D}$, is then passed through two MLP heads for scene segmentation and feature encoding, as detailed in~\cite{Engelcke2021GENESISV2IU}, where $D$ represents the dimension of the feature map. Specifically, based on $\zeta \in \mathbb{R}^{N\times D}$ and the two respective MLPs, we obtain the encodings $\zeta^{seg} \in \mathbb{R}^{N\times D}$ and $\zeta^{feat} \in \mathbb{R}^{N\times D}$. The encoding $\zeta^{seg} \in \mathbb{R}^{N\times D}$ is used for scene segmentation (Section~\ref{subsec: Scene Segmentation}), while $\zeta^{feat} \in \mathbb{R}^{N\times D}$ is directly used for scene reconstruction  (Section~\ref{subsec:Scene Reconstruction}). This preprocessing step adeptly transforms the input into these two embeddings in a noise-robust manner, which are subsequently utilised by the rest of the model architecture.

\subsection{Scene Segmentation}
\label{subsec: Scene Segmentation}
Given the point-wise embedding $\zeta^{seg}$, we apply an instance colouring stick-breaking process~(IC-SBP)~\cite{Engelcke2021GENESISV2IU} for scene segmentation. IC-SBP is a clustering algorithm that predicts $K$ soft attention masks $\{\mathbf{m}_{k}\}_{k=1}^{K}\in \left [ 0,1 \right ]^{N\times 1}$ that follow a stick-breaking process. The predicted attention masks are randomly ordered, which is achieved by stochastically sampling cluster seeds from the point-wise embeddings. We denote the first attention mask, $\mathbf{m}_{0}$, as the background mask and the last attention mask, $\mathbf{m}_{K}$, as the redundant scope that tracks the unexplained pixels. We refer readers to~\cite{Engelcke2021GENESISV2IU,Wu2022ObPoseLC} for the implementation details. The observed scene point cloud can then be segmented into several object point clouds denoted by $\mathbf{P}_{k}^{obs}$ where $k$ is the object index. 

\subsection{Pose Estimation with Shape Completion} 
\label{subsec:Pose Estimation with Shape Completion}
A 6D pose for each segmented object can be estimated by computing a minimum volume bounding box over the object's masked point cloud~\cite{Wu2022ObPoseLC}. However, estimating a full 6D pose for a partially observed object is challenging and prone to inaccuracies and high variance when estimated from only the information available from a single view. To overcome this challenge, we propose a shape completion module that estimates the shape of occluded parts of objects.

First, each observed object point cloud, $\mathbf{P}_{k}^{obs}$, is transformed into a canonical pose, $\mathbf{P}_{k}^{can}$, by transforming each point in the point cloud as follows (in the same manner as~\cite{Wu2022ObPoseLC}):
\vspace{-0.3cm}
\begin{equation}
    \mathbf{P}_{k}^{can} = \frac{2}{b}(\mathbf{R}_{k})^{-1}\left (\mathbf{P}_{k}^{obs} - \mathbf{T}_{k} \right )
\label{eq:transform}
\end{equation}

where $\mathbf{T}_{k}$ is the location of the sampled seed of the object mask $\mathbf{m}_k$ in the IC-SBP algorithm and $\mathbf{R}_{k}$ is the rotation matrix. For the first transformation, $\mathbf{R}_{k}$ is set to the identity matrix. The bounding box size $b\in \mathbb{R}^+$ is initialised to a sufficiently large number for all objects. 

For each object, the shape completion module encodes the transformed point cloud $\mathbf{P}_{k}^{can}$ and the masked scene embedding, $\zeta_{k}^{feat} = \zeta^{feat}\odot\mathbf{m}_{k}$, using a KPConv-based autoencoder into a shape embedding $\mathbf{e}_{k}$. Using a tri-plane-based GRAF~\cite{mildenhall2020nerf,schwarz2020graf,niemeyer2021giraffe,chan2022efficient}, which we denote as $\pi^{shape}$, each shape embedding, $\mathbf{e}_{k}$, is decoded into a voxelised representation which is used for shape completion. A GRAF is a generative NeRF that models the 3D geometry and texture of an object by predicting the occupancy and the colour of any given query point $\mathbf{p}$. Specifically, $\pi^{shape}$ learns to map the latent encoding $\mathbf{e}_{k}$ to the occupancy logits, 
\begin{equation}
   \sigma_{k}(\mathbf{p}) = \pi^{shape}(\mathbf{p},\mathbf{e}_{k}(\zeta_{k}^{feat})), 
\end{equation}
 at the given query point $\mathbf{p}$. The colour prediction from the original GRAF formulation is discarded here as only occupancy is needed for shape completion, resulting in the shape-GRAF, $\pi^{shape}$. 

To reduce the computational overhead of approximating the 3D shape of the object, we divide each object bounding box into $\mathbf{S}$ voxels along each dimension. The centre position of each voxel, $\mathbf{p}_{v,k}$, is evaluated with an occupancy probability scalar value computed as in~\cite{Wu2022ObPoseLC}:

\begin{equation}
    \mathbf{\phi}_{k}(\mathbf{p}_{v,k}) = \mathrm{tanh}\left ( \mathrm{softplus}\left ( \mathbf{\sigma}_{k}(\mathbf{p}_{v,k}) \right ) \right ).
\end{equation}
Voxels are considered occupied if $\mathbf{\phi }(\mathbf{p}_{v,k})>\mathbf{\phi}_{T}$, with $\mathbf{\phi}_{T}$ acting as a threshold. 
Occupied voxels are used as the shape completion points, $\mathbf{P}_{k}^{comp}$. A minimum volume bounding box that contains both $\mathbf{P}_{k}^{comp}$ and $\mathbf{P}_{k}^{obs}$ is used to represent the object pose. Again, the object point cloud, $\mathbf{P}_{k}^{obs}$, is transformed to a canonical pose, $\mathbf{P}_{k}^{can}$, using the improved object pose estimate and~\cref{eq:transform}, where the improved estimate of the object's position is $\mathbf{T}_{k}$ and the rotation is $\mathbf{R}_{k}$.

\subsection{Scene Reconstruction}
\label{subsec: Scene Reconstruction}
A latent embedding $\mathbf{z}_{k}$ is created by encoding the updated $\mathbf{P}_{k}^{can}$ and $\zeta_{k}^{feat}$ using a KPConv-based encoder. $\mathbf{z}_{k}$ is parameterised as a Gaussian and then used by a tri-plane-based GRAF to decode the 3D shape and colour of each object. We denote the GRAF that decodes the object embedding $\mathbf{z}_k$ as the object-GRAF. We follow the method of \cite{chan2022efficient} to predict colour and occupancy logits for each object given the object embedding. 

To reconstruct the entire scene each object must be reconstructed individually along with the background component. The occupancy and colour of the background are reconstructed by first encoding the observed point cloud together with the masked scene embedding, $\mathbf{\zeta}_{0}^{feat} = \mathbf{\zeta}^{feat}\odot\mathbf{m}_{0}$, using a KPConv encoder to predict a background latent embedding $\mathbf{z}_{bg}$ before being decoded by a third tri-plane-based background-GRAF.



Computing the occupancy probability of the entire scene, $\mathbf{\phi}_{ \textup{scene}}(\mathbf{p})$, at the queried points $\mathbf{p}$ given the viewing direction, $\mathbf{d}$, using the outputs of the object and background GRAFs can be completed as follows~\cite{Wu2022ObPoseLC}:

\begin{equation}
\scalebox{0.8}{$
    \mathbf{\phi}_{ \textup{scene}}(\mathbf{p}) = \mathrm{tanh}\left ( \sum_{k=0}^{K-1}\mathrm{softplus}\left (\mathbf{\sigma}_{k}(\mathbf{p}) \right ) \right )$}
\label{eq:apex_norm}
\end{equation}

For the colours of the scene $\mathbf{c}_{\textup{scene}}(\mathbf{p},\mathbf{d})$, we can compute the weighted mean~\cite{Wu2022ObPoseLC,Stelzner2021Decomposing3S}:
\begin{equation}
\scalebox{0.8}{$
    \mathbf{c}_{\textup{scene}}(\mathbf{p},\mathbf{d}) = \sum_{0}^{K-1}\hat{\mathbf{\phi}}_{k}(\mathbf{p})\mathbf{c}_{k}(\mathbf{p},\mathbf{d}),$}
\end{equation}
with $\hat{\mathbf{\phi}}_{k}(\mathbf{p}) = \mathbf{\phi}_{\textup{scene}}(\mathbf{p})\underset{K}{\mathrm{softmax}}\left ( \mathbf{\sigma}_{k}(\mathbf{p}) \right ) $. 

\subsection{Training}

With known depth, the object-GRAF, shape-GRAF and background-GRAF only require two evaluations in each training iteration~\cite{Stelzner2021Decomposing3S}, i.e. one evaluation at the surface and one evaluation at points sampled between the camera and the surface. We denote these two points as $\mathbf{p}_{surf}$ and $\mathbf{p}_{empty}$ respectively. We refer readers to~\cite{Stelzner2021Decomposing3S} for how $\mathbf{p}_{surf}$ and $\mathbf{p}_{empty}$ are sampled with the known depth. The object-GRAF and the background-GRAF are queried at $\mathbf{p}_{surf}$ to learn the reconstruction of the geometry and the colour of the scene by maximising the likelihood of the observations:



\begin{equation}
\scalebox{0.62}{$\mathcal{L}_{\text{obs}} = -\log \Bigg(\mathbf{\phi}_{\textup{scene}}(\mathbf{p}_{surf})\Bigg(\sum_{k=0}^{K-1} \Big(\mathcal{N}(\mathbf{x}|\mathbf{c}_{k}(\mathbf{p}_{surf},\mathbf{d}_{surf}),\sigma_{\text{std}}^{2}) \odot \hat{\mathbf{\phi}}_{k}(\mathbf{p}_{surf})\Big)\Bigg)^{\eta} \Bigg)$}
\end{equation}


with $\eta \sim  Ber(\mathbf{\phi}_{\textup{scene}}(\mathbf{p}_{surf}))$ and $\mathbf{d}_{surf}$ being the view directions of points $\mathbf{p}_{surf}$. To prevent the background module from overfitting to the scene, we use RANSAC~\cite{1981Random} to detect the largest plane, e.g. the table, in the scene and constraint the background module to only learn to reconstruct the detected plane as the background.
The GRAFs are queried at $\mathbf{p}_{empty}$ to learn to fit the empty space of the scene. However, $\mathbf{p}_{empty}$ can be sampled at places that are known to be empty, e.g. the points outside of the object bounding boxes. Instead, for the object-GRAFs, we choose the voxel centres of the object bounding boxes $\mathbf{p}_{v,k,\textup{obj}}$, and find those in the empty space as the sampled points denoted as $\mathbf{p}_{e,k,obj}$. The points $\mathbf{p}_{e,k,obj}$ are determined using the known camera parameters and the depth observations. We thus only evaluate the background-GRAF at $\mathbf{p}_{empty}$ and minimise the occupancy in the empty space by computing the $\mathcal{L}_{\text{empty}}$ as follows:
\begin{equation}
\scalebox{0.9}{$    \mathcal{L}_{\text{empty}} = \mathbf{\phi}_{0}(\mathbf{p}_{empty})/\mathbf{\rho }_{empty} -\sum_{k=1}^{K-1}\log(1-\mathbf{\phi}_{k}(\mathbf{p}_{e,k,obj}))$}
\end{equation}
where $\mathbf{\rho }_{empty}$ is the probability density of the point $\mathbf{p}_{empty}$ being sampled for the background-GRAF. To further improve the reconstruction accuracy, we sample $N_{r}$ points of $\mathbf{p}_{empty}$ and $\mathbf{p}_{surf}$ from multi-view observations in each training iteration. Note that this is only used for training, and at test time, the inference only uses single-view input.
To supervise the shape-GRAF, $\pi^{shape}$, we use a self-distillation loss to save computation by reusing the evaluations of the object-GRAF. Concretely, we compute the trilinear interpolation $T_{tr}(\mathbf{p}_{v,k},\mathbf{\phi}(\mathbf{p}_{v,k,\textup{obj}}))$ of the occupancy predictions of the object-GRAF at $\mathbf{p}_{v,k}$. We then distil the shape information from the object-GRAF to the shape-GRAF by minimising the KL divergence between the Bernoulli distribution of $T_{tr}(\mathbf{p}_{v,k},\mathbf{\phi}(\mathbf{p}_{v,k,\textup{obj}}))$ and $\mathbf{\phi}(\mathbf{p}_{v,k})$:
\begin{equation}
\scalebox{0.8}{$
\mathcal{L}_{\text{shape}} = \sum_{k=0}^{K-1} \mathbb{KL}(Ber(T_{tr}(\mathbf{p}_{v,k},\mathbf{\phi}(\mathbf{p}_{v,k,\textup{obj}})))||Ber(\mathbf{\phi}(\mathbf{p}_{v,k})))$}
\end{equation}
The IC-SBP attention masks are supervised via a L2 Loss:
\begin{equation}
\label{eq:att}
   \scalebox{0.8}{$ \mathcal{L}_{\text{att}}= \sum_{k=0}^{K-1} (\mathbf{m}_{k} - \hat{\mathbf{\phi}}_{k}(\mathbf{p}_{surf}))^{2} + (\mathbf{m}_{K} - (1-\mathbf{\phi}_{\textup{scene}}(\mathbf{p}_{surf})))^{2}$}
\end{equation}
Here, $\mathbf{m}_{K}$ is the redundant scope. The last term in \cref{eq:att} encourages the redundant scope to treat points that have a low observation likelihood as not explained points. We also apply the sparsity loss $\mathcal{L}_{\text{sparsity}}$ introduced in \cite{yu2021plenoctrees} to encourage the model to choose empty space when no observations are available, e.g. the space beneath the table in the table-top scene. The total loss is computed as an aggregation of all the aforementioned losses: $\mathcal{L} =  \mathcal{L}_{\text{empty}} + \mathcal{L}_{\text{obs}} + \mathcal{L}_{\text{shape}} + \mathcal{L}_{\text{att}} + \mathcal{L}_{\text{sparsity}}$.

\section{Experiments}
\label{Results}

In this section, we describe the experimental setup and hardware employed in our experiments, the performance metrics used to evaluate the effectiveness of our model alongside baselines, and present the primary outcomes of our study. Specifically, we investigate the following questions: (i) Does \model~achieve improved performance for scene understanding compared to NeRFs, especially in terms of inference time and reconstruction accuracy? (ii) Compared to pre-trained object-centric features, can the object-centric latent representation improve performance on downstream tasks such as object matching? and, (iii) Does \model~ with shape completion have better pose estimation compared to other unsupervised, object-centric pose estimation methods in real-world scenarios?

\subsection{Experimental Setup}

\begin{figure}[hbtp!]
  \centering
  \includegraphics[width=0.7\linewidth]{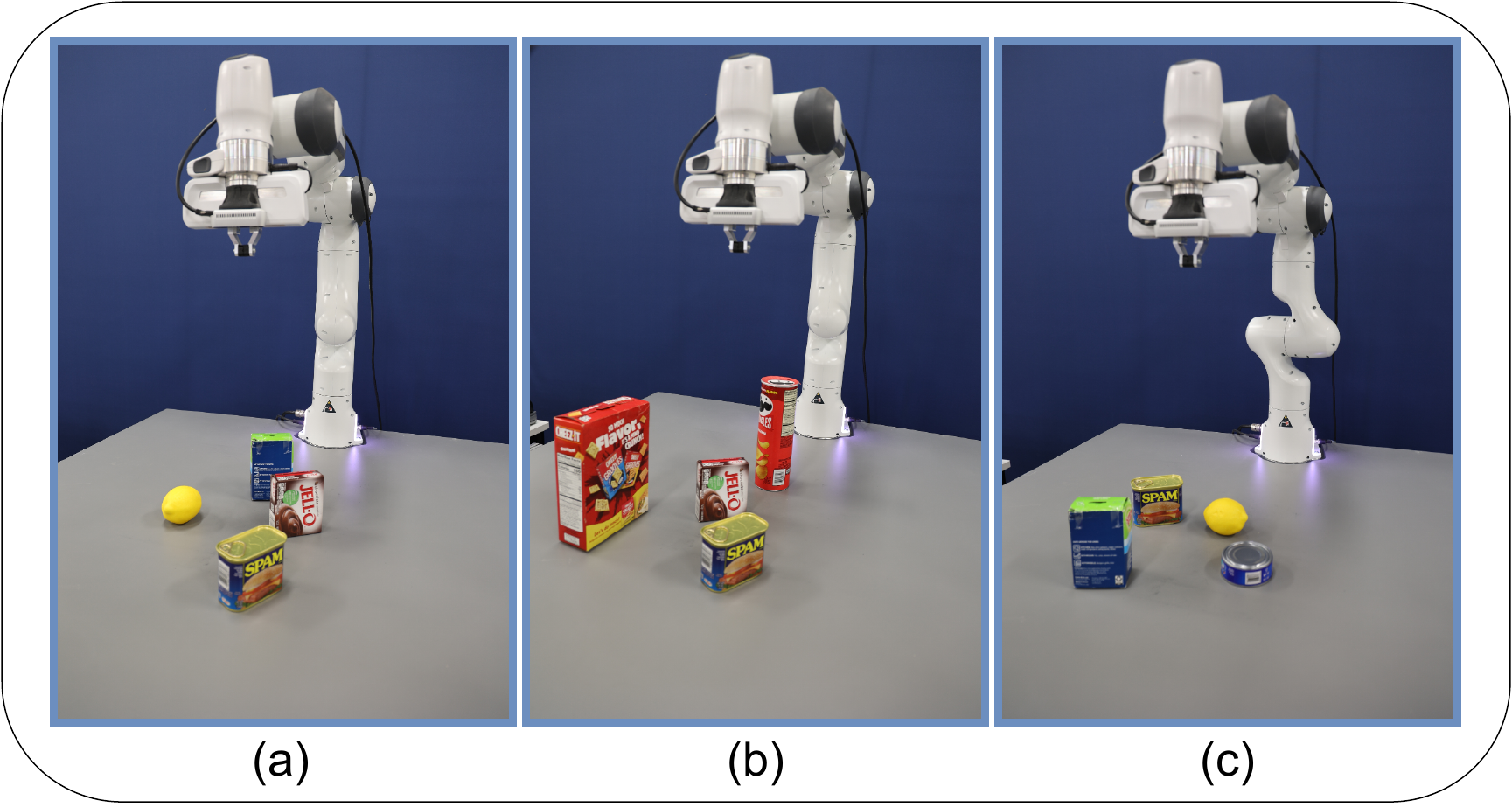}
  \caption{Example configurations used in the experiments. Panels depict the 7-DoF Franka Panda robot together with YCB objects randomly selected and configured on the tabletop.}
  \label{fig:example_configuration}
  \vspace{-0.2cm}
\end{figure}

Data collection and real-world testing of \model~utilised a 7-DoF Franka Panda manipulator equipped with an Intel RealSense camera. The system also included a computer with an NVIDIA GeForce RTX 3090 GPU. A subset of YCB objects, including the mustard bottle, fish can, banana, apple, lemon, orange, potted meat, cracker box, pringles, chocolate pudding box, soup can, sponge, gelatin box and spray bottle, were employed for experiments. Each scene collected in the training set consisted of two to four objects randomly arranged on the table (see Figure~\ref{fig:example_configuration}). The objects were arranged in graspable poses with inter-objects occlusions. For model training, 200 scenes were collected with each scene captured from four viewpoints, providing camera extrinsics and RGB-D images as model inputs. For assessing 3D reconstruction performance from a single viewpoint, a test set of scenes following the same configuration as the training set were collected. Training of the model took approximately 3 days on a single NVIDIA RTX A6000 GPU.

\subsection{Scene Reconstruction}
\label{subsec:Scene Reconstruction}

In this section, we assess the scene reconstruction capabilities of \model~compared to a recent state-of-the-art baseline. Additionally, we demonstrate \model's capacity to \textit{imagine} missing parts in the scene in an object-centric fashion based on only a single-view.

\begin{figure*}[tb!]
  \centering
  \includegraphics[width=0.6\textwidth]{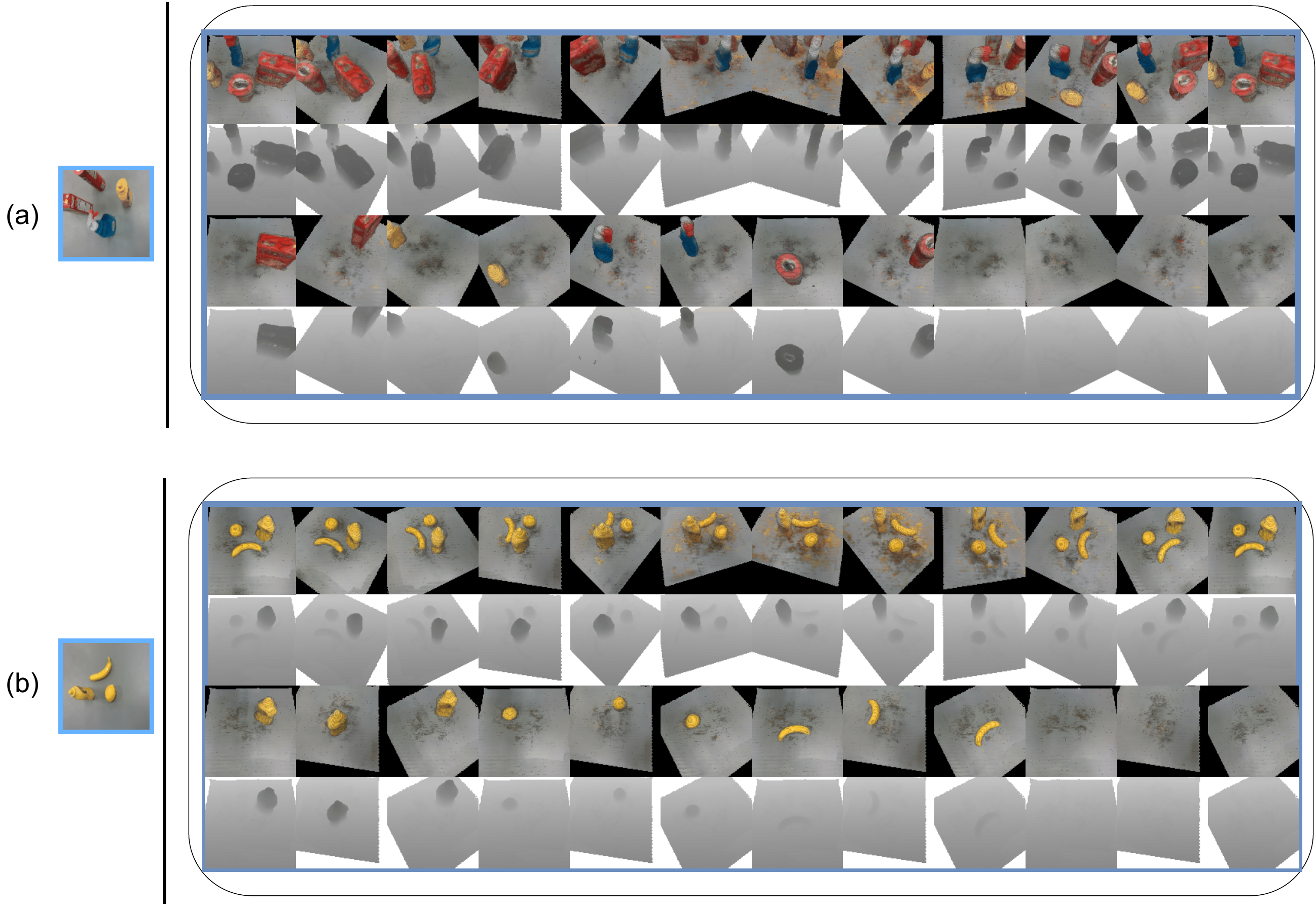}
  \caption{Given a single view of the scene, \model~produces full scene reconstructions from arbitrary vantage points. For two examples, (a) and (b), the top two rows show the scene reconstructions for RGB and depth respectively from various viewpoints. The following two rows show RGB and depth reconstructions for individual objects and the background components. Example (a) demonstrates the ability of \model~to reconstruct the shapes of the cracker box and the the chips despite being partially out of view in the input image.} 
  \label{fig:\model_multiview_reconstructions}
  \vspace{-0.4cm}
\end{figure*}

\textbf{Metric.} 
To evaluate the 3D reconstruction capabilities of our model, we use the directed Hausdorff distance (DHD) as our metric.
\begin{equation}
    d_{H}(A,B) = \sum_{a \in A} \min_{b \in B} \|a - b\|.
\end{equation}
In our case, $A$ is the observed ground truth point set, $B$ is the reconstructed point set, and the $\|\cdot\|$ represents the Euclidean distance. Note that, the directed Hausdorff distance only performs the calculation based on the points from the original ground truth point set as reference. This is different from the Chamfer distance, a metric commonly used to measure the dissimilarity between two sets of points. We use the directed Hausdorff distance instead of the Chamfer distance as the point set $B$ generated by \model~can contain \textit{imagined} points that do not exist in set $A$ (see Fig.~\ref{fig:\model_multiview_reconstructions}). The absence of reference points in the original set prevents the comparison of minimum distances for these generated points using the Chamfer distance. 

\textbf{Baselines.} We use \textit{depth-nerfacto} from \textit{nerfstudio}~\cite{nerfstudio} as one of our baselines. According to the official documentation, the model is not based on any single existing published work. Instead, it integrates numerous published methods that have been identified as highly effective for reconstructing real data when used jointly. In particular, the model combines camera pose refinement, per-image appearance conditioning, proposal sampling, scene contraction, and hash encoding to accelerate training. 
In terms of 3D reconstruction, we compare our model to \textit{depth-nerfacto} using several different configurations. We provide the NeRF with 8, 16, and 32 views for reconstruction and optimize it for between 5000 to 15000 training iterations. We also compare our model to another OCGM, ObSuRF~\cite{Stelzner2021Decomposing3S}. ObSuRF also employs generalizable NeRFs as the object decoder but encodes the spatial information and object appearance into a single latent embedding, preventing the estimation of a 6D pose.
We compare the directed Hausdorff distance between ground truth and reconstructions across 12 test scenes, assessing the time taken by each model to reconstruct a given scene. The 12 test scenes are collected in the same way as the training scenes, although, with unseen, random configurations.

\textbf{Reconstruction Results.} \model~only needs to be trained once and does not necessitate retraining when the tabletop configuration is altered. In contrast, \textit{depth-nerfacto} lacks the ability to generalise and requires retraining for each modified scene configuration. Additionally, \model~can reconstruct the scene from a single-view RGB-D image and \textit{imagine} the 3D shapes, while \textit{depth-nerfacto} requires multiple views as inputs for each new scene. The results are presented in Table~\ref{tab:reconstruction_results}\footnote{The NeRFs are off-the-shelf models without any modifications, we observe that providing more views or additional training time does not necessarily improve performance. This indicates that training for 5000 iterations is sufficient.}. 
It is evident that \model~achieves significantly faster test time inference, taking only a fraction of a second, whilst also exhibiting superior reconstruction performance, surpassing the performance of the baseline method by an order of magnitude. This improved test-time efficiency of \model~satisfies the outlined desiderata for real-time robot operations not exhibited by NeRFs. We observe that ObSuRF, another OCGM, also elicits fast test time inference, however, it does not perform as well as \model~ in reconstruction accuracy.   

\begin{table}[htbp!]
  \centering
  \caption{3D reconstruction results for \model~vs NeRF and ObSuRF baselines. \model~outperforms the NeRF baselines with up to 32 views and 15000 fitting steps both in terms of speed and reconstruction error (distance to ground-truth). The results for foreground reconstruction is summarised in \model~(FG only). \model~also outperforms ObSuRF in terms of reconstruction accuracy.}
  \label{tab:reconstruction_results}
  \scalebox{0.7}{
  \begin{tabular}{ccccc}
    \toprule
    Model & Views $\downarrow$ & Fitting steps & Test Time (s) $\downarrow$ & DHD $\downarrow$ \\
    \midrule
    \model & 1 & 0 & $\mathbf{0.24 \pm 0.01}$ & $\mathbf{0.0075 \pm 0.0010}$\\
    \model~(FG only) & 1 & 0 & $\mathbf{0.23 \pm 0.01}$ & $0.0081 \pm 0.0023$\\
    ObSuRF & 1 & 0 & $\mathbf{0.01 \pm 0.01}$ & $0.0130 \pm 0.0060$\\\midrule
    NeRF & 8 & 5000 & $127.41 \pm 1.11$ & $0.0351 \pm 0.0086$ \\
    NeRF & 16 & 5000 & $129.12 \pm 1.31$ & $0.0384 \pm 0.0077$ \\
    NeRF & 32 & 5000 & $128.02 \pm 1.34$ & $0.0423 \pm 0.0057$ \\
    NeRF & 32 & 10000 & $245.73 \pm 2.11$ & $0.0463 \pm 0.0079$ \\
    NeRF & 32 & 15000 & $364.83 \pm 1.33$ & $0.0501 \pm 0.0109$ \\
    
    \bottomrule
  \end{tabular}}
  \vspace{-0.3cm}
\end{table}

\begin{figure}[htbp!]
\centering
\includegraphics[width=0.7\linewidth]{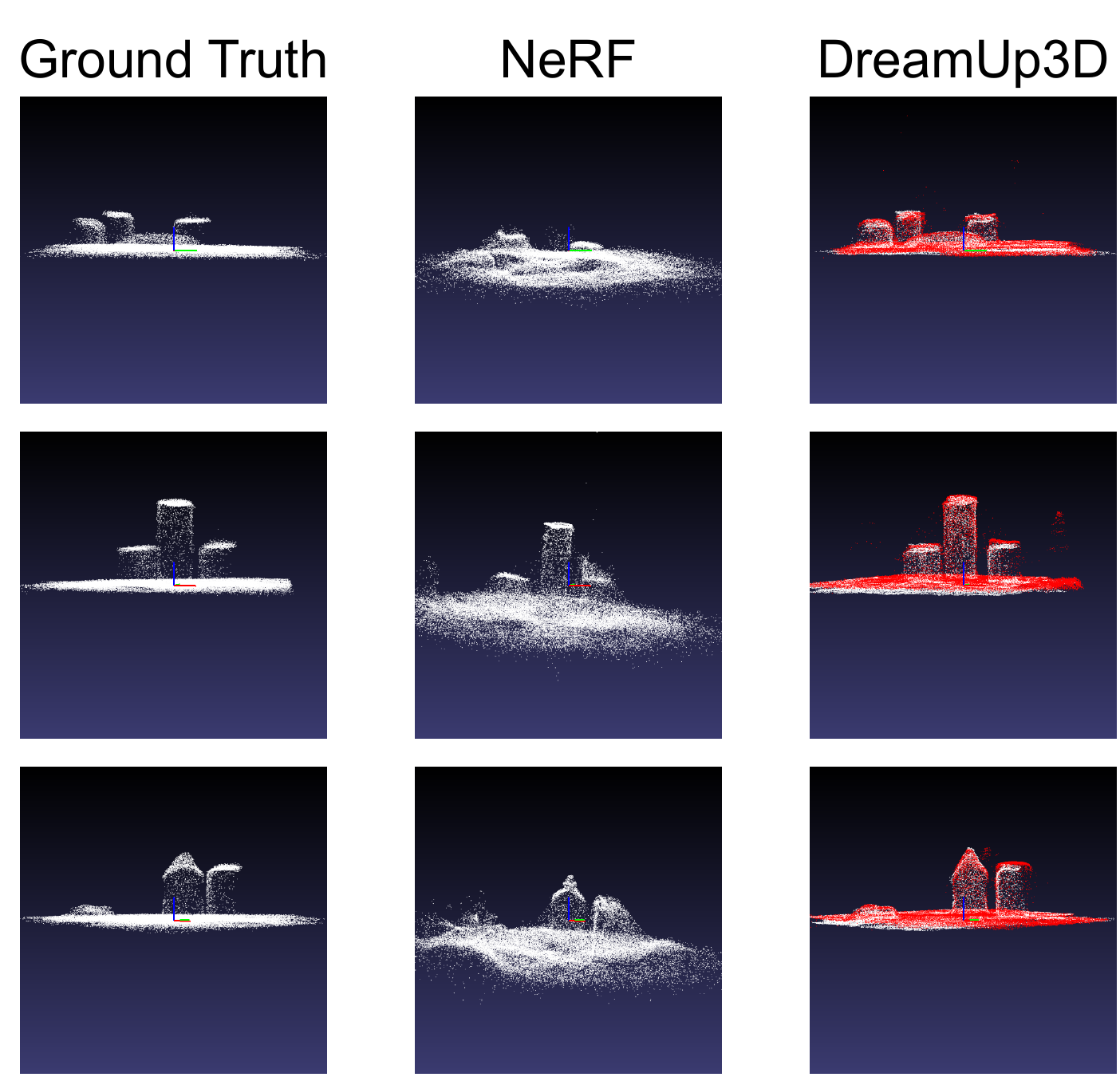}
\caption{Comparison of point cloud ground truth, NeRF reconstruction using 32 views, and \model~reconstruction using a single view image as input. In the case of \model~we superimpose the ground truth (white) to the reconstruction (red). All images are taken from the same view point with the size of the point clouds increased for visual clarity.}
\label{fig:pointcloud}
\vspace{-0.7cm}
\end{figure}

Example RGB and depth reconstructions are depicted in Figure~\ref{fig:\model_multiview_reconstructions}. Note that in the case of Figure~\ref{fig:\model_multiview_reconstructions} (a), it can be observed that part of the objects in the input image are out of frame. \model~demonstrates its ability to \textit{imagine} the missing parts of the objects, this characteristic can be attributed to the object-centric scene factorisation of the model. Concretely, \model~first recognises the partially observed objects in the scene using the learned object encoder and reconstructs the full shape via the learned object decoder. Also, in Figure~\ref{fig:pointcloud}, we compare the ground truth point cloud to those reconstructed by \textit{depth-nerfacto} and \model. All images are taken from the same camera view. It is evident that \model~captures the underlying point cloud geometry and objects with greater precision (see Table~\ref{tab:reconstruction_results}). ObSuRF encodes spatial information and appearance information of the object into a single latent embedding, which poses a higher requirement on the learning capacity. We thus find that despite ObSuRF's ability to reconstruct the scene, it fails to segment the scene into meaningful objects, leading to strong reconstruction accuracy although without sufficient scene understanding. 

\subsection{Object-centric Representation}
We consider an object matching task to evaluate the real-world applicability of the learned object-centric representations.

\textbf{Metric.} For the object matching task, we report the matching accuracy as our metric.

\textbf{Baselines.} We first conduct an evaluation of our model by comparing it with recent visual-semantic matching techniques, namely CLIPSEMFEAT-N~\cite{Goodwin2021SemanticallyGO}, which utilises the pre-trained CLIP model~\cite{Radford2021LearningTV} for object matching. Following \cite{Radford2021LearningTV}, an instance segmentation network~\cite{xiang2021learning} first extracts crops of all the objects present in both the current scene and the given goal image. Cropped objects are then encoded into visual features using the CLIP model. A classification matrix is constructed as the dot product between the normalised visual features and the semantic features acquired by encoding the known class names into a text embedding using the text encoder of the CLIP model. For~\model, we directly compute the L2-distance and the cosine distance (dot product) of the normalised object-centric latent representations for object matching. We evaluate our model and baselines on a tabletop scene involving multiple objects. For each experiment, we randomly select two to four objects from the YCB dataset, with evaluations conducted across 20 unique scenes in the real-world. 

\textbf{Object Matching Results.} 
The quantitative results for our object-matching experiment are summarised in Table \ref{tab:matching_results}. We observed that the CLIPSEMFEAT-N approach fails to perform object matching well in our experiment as the CLIPSEMFEAT-N approach requires semantically distinguishable visual inputs. However, objects such as the potted meat can only expose a metal surface from the fixed top-down view, which is required by the GG-CNN grasping planner~\cite{Morrison2018ClosingTL} used in \cite{Goodwin2021SemanticallyGO}. This partial observation leads to the low accuracy of the CLIPSEMFEAT-N approach. \model, on the other hand, performs object matching using the object-centric latent representation learned from the scene reconstruction. Our approach thus does not require known object classes and allows for fully self-supervised learning. Using the learned object latent representation \model~achieves a significantly improved matching accuracy compared to CLIPSEMFEAT-N. However, we also find that matching among objects that share similar shapes, e.g. the chocolate pudding box and the strawberry gelatin box, can be erroneous, which indicates the limitation of performing object matching using latent representations learned only from scene reconstruction. As an ablation, we include the accuracy of the L2-distance versus the cosine distance (dot product) for the object matching and see a notable improvement when using the cosine distance.
\vspace{-0.2cm}
\begin{table}[ht]
  \centering
  \caption{Object matching results \model~vs CLIPSEMFEAT-N baseline.}
  \label{tab:matching_results}
  \scalebox{0.9}{
  \begin{tabular}{cc}
    \toprule
    Model & Accuracy [\%] \\
    \midrule
    \model~(cosine distance) & 57.6 \\
    \model~(L2-distance) & 55.9\\
    \midrule
    CLIPSEMFEAT-N &  27.1\\
    \bottomrule
  \end{tabular}}
\vspace{-0.5cm}
\end{table}

\subsection{Unsupervised Pose Estimation}
In this section, we assess the unsupervised pose estimation capabilities of DreamUp3D compared to a recent state-of-the-art baseline, ObPose~\cite{Wu2022ObPoseLC}. We also analyse the common failure modes of shape-based pose estimation caused by the partial observation of the scene.

\textbf{Metric.} 
Due to the symmetry of some objects, it is possible that there is no unique correct orientation, we thus evaluate the pose using the mean intersection over union (mIoU) accuracy of the predicted bounding boxes and the ground-truth bounding boxes with various thresholds~\cite{tosi2020distilled,you2020keypointnet} as the metric.

\textbf{Baseline.} The pose estimation performance of \model~is evaluated compared to the ObPose model~\cite{Wu2022ObPoseLC}. ObPose is the first OCGM that performs unsupervised 6D pose estimation but without the use of a shape completion module to predict the full object shape.

\begin{figure}[tbp!]
\centering
\includegraphics[width=0.98\linewidth]{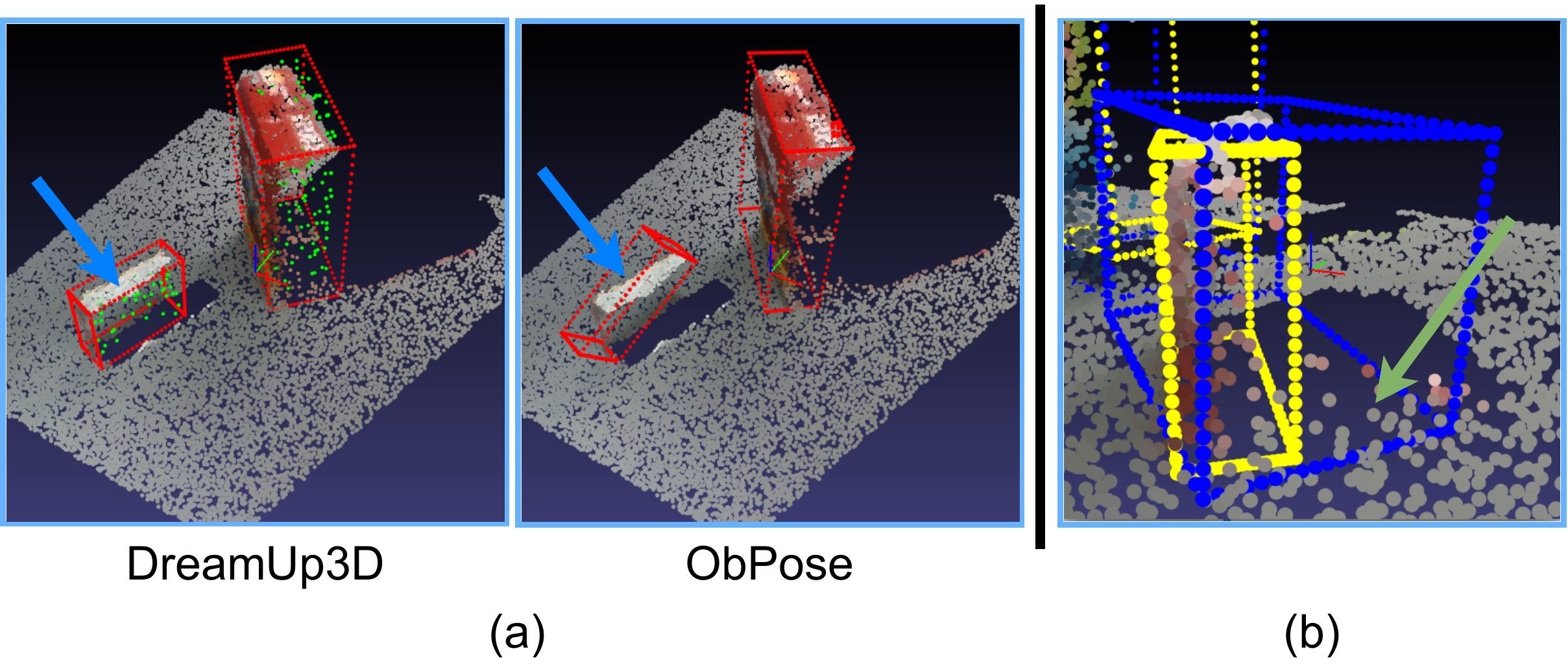} 
\caption{Object pose estimation in the real world. (a): The pose estimation with shape completion (\model) and without shape completion (ObPose). The predicted object point clouds are visualised using green points. (b) The noisy point clouds projected from the depth observation at the edges of the objects, which become outliers for the minimum bounding box.}
\label{fig:pose}
\vspace{-0.3cm}
\end{figure}

\textbf{Pose Estimation Results.} The quantitative results for pose estimation performance comparing ObPose~\cite{Wu2022ObPoseLC} to \model~ are summarised in Table \ref{tab:pose_results}. We observe two main failure modes of ObPose, occlusions and noisy scene observations, and demonstrate these in Figure~\ref{fig:pose}. With occlusion, the observed object point clouds are not sufficient to accurately approximate the full object shape, which thus leads to an erroneously oriented minimum bounding box (see Figure~\ref{fig:pose} (a)). The noisy scene point clouds are caused by the accuracy-drop of the depth sensor at the object edges. Part of the object point clouds can be mistakenly projected to the table surface and therefore become outliers for the minimum bounding box (see Figure~\ref{fig:pose} (b)). To alleviate this problem, we randomly drop part of the observed point clouds to reduce the outliers and preserve the in-painted point clouds generated from the shape completion module. We observe a clear improvement as the observed object point clouds reduce (see Table \ref{tab:pose_results}). This indicates the proposed shape completion module can provide more robust shape information compared to the ObPose baseline.
\begin{table}[ht]
  \centering
  \caption{Pose estimation results \model~vs ObPose baseline. The mean IoU (mIoU) accuracy with thresholds of $0.3,0.5$ and $0.7$ are reported. The pose estimation using less observed (obs.) point clouds (pcds) with shape completion (SC) achieves improved performance.}
  \label{tab:pose_results}
  \scalebox{0.75}{
  \begin{tabular}{ccccc}
    \toprule
    Model & mIoU (0.3) $\uparrow$  & mIoU (0.5) $\uparrow$  & mIoU (0.7) $\uparrow$ \\
    \midrule
    \model~(SC only) & \textbf{0.97} & \textbf{0.75} & 0.15\\
    \model~(SC + 10\% obs. pcds) & 0.95 & 0.75 & \textbf{0.17} \\
    \model~(SC + 30\% obs. pcds) & 0.90 & 0.69 & \textbf{0.17} \\
    \model~(SC + 50\% obs. pcds) & 0.92 & 0.68 & 0.14 \\
    \model~(SC + 70\% obs. pcds) & 0.88 & 0.68 & 0.08 \\
    \midrule
    ObPose~(obs. pcds only) & 0.90 & 0.61 & 0.03\\
    \bottomrule
  \end{tabular}}
  \vspace{-0.3cm}
\end{table}

\begin{figure}[htbp!]
\centering
\includegraphics[width=0.98\linewidth]{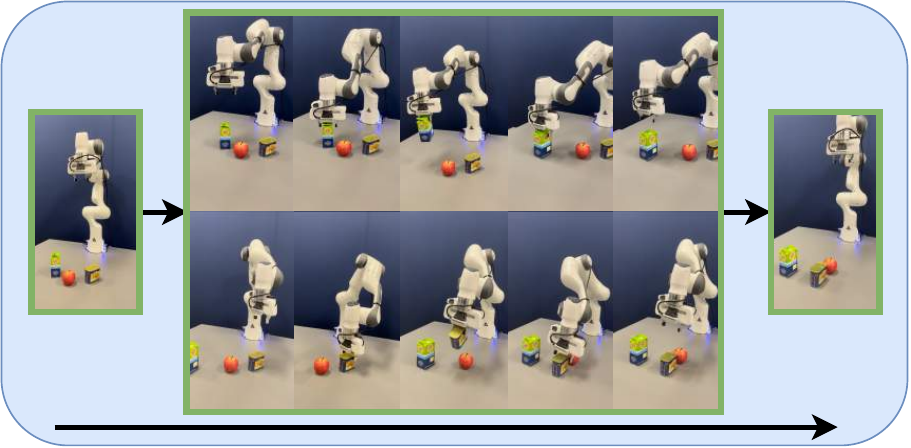} 
\caption{Object manipulation in the real world. The agent detects and moves the objects to their canonical poses using the inferred object poses. The robot arm moves the blue and green box away from the apple (first row, left to right). The robot arm moves the can from behind the apple to the front (second row, left to right).}
\label{fig:rearrangement}
\vspace{-0.6cm}
\end{figure}

Leveraging the estimated object pose, we additionally demonstrate \model’s ability to flexibly rearrange real-world objects, with the use of object position and orientation, in Figure \ref{fig:rearrangement}. 

\subsection{Limitations}
Although our method has the advantage of providing strong scene reconstructions, latent object embeddings and 6D pose estimation, the reconstruction could see significant improvements using an RGB-D sensor with greater accuracy. It is important to note that due to sensor limitations, point clouds captured from different angles do not perfectly overlap. This lack of alignment can partly be attributed to errors in camera pose estimation, which are influenced by the robot's forward kinematic accuracy and camera calibration accuracy. Additionally, depth errors can significantly impact the accuracy of our reconstructions, this typically occurs when scenes include shiny surfaces like metals and reflective plastics. These factors collectively contribute to the reconstruction error. 

There is potential for \model~ to generalise to out-of-distribution cases where scenes feature a greater number of objects than those seen in training thanks to the the IC-SBP algorithm and the object-centric learning paradigm. We test this hypothesis by collecting additional highly clustered scenes with 4-6 objects including scenes with unseen objects not included in our training set. As expected, \model~ does not generalise to scenes with unseen objects, achieving a DHD of $0.0105 \pm 0.0039$. This is a known limitation of OCGMs, as the encoder-based approach of single-view 3D reconstruction is essentially a recognition model~\cite{tatarchenko2019single}. Note that reconstructing unseen objects from single-view observation is a highly ill-posed task. Generalisation to scenes with a greater number of objects is also limited (DHD of $0.0090 \pm 0.0021$) likely caused by the closeness of objects within the test scenes. Both of these limitations could potentially be addressed by scaling up both data and computational resources to the model.

Finally, our model and experiments focus specifically on table-top scenes due to the RANSAC plan detection. Extending our model to more complex scenes in the wild requires the adaptation of the model and specifically the background modelling component. The hyperparameters such as the kernel sizes of the KPConv encoder and the GRAF network sizes are environment-dependent. They need to be chosen according to the sizes and the complexity of the scene.

\section{Conclusion}
\label{Conclusion}

We propose \model~as an efficient and robust approach for performing 3D object-centric scene inference, object-level representation learning, and 6D pose estimation. In contrast to related work, \model~achieves these tasks without the need for retraining on new scenes at test time and without requiring multiple views of a static scene. This makes \model~more readily suited to robotics tasks. Compared to recent baselines, we demonstrate \model's improved reconstruction quality and its ability to \textit{imagine} occluded or missing parts of objects in the input image.

Further work is needed to pursue reconstructions in challenging scenarios involving reflective surfaces, although acquiring accurate depth estimates in these conditions using RGB-D cameras is a well-known problem. In the context of object manipulation, incorporating 3D reconstruction for improved grasping presents a promising avenue for future research. 
\section*{Acknowledgment}
This work was supported by the EPSRC Programme Grant (EP/V000748/1), an Amazon Research Award and the China Scholarship Council. The authors would like to acknowledge the use of the University of Oxford Advanced Research Computing (ARC) facility in carrying out this work. \url{http://dx.doi.org/10.5281/zenodo.22558}. 

\ifCLASSOPTIONcaptionsoff
  \newpage
\fi



\bibliographystyle{IEEEtran}
\bibliography{IEEEabrv,bib}

\begin{thebibliography}{10}
\providecommand{\url}[1]{#1}
\csname url@rmstyle\endcsname
\providecommand{\newblock}{\relax}
\providecommand{\bibinfo}[2]{#2}
\providecommand\BIBentrySTDinterwordspacing{\spaceskip=0pt\relax}
\providecommand\BIBentryALTinterwordstretchfactor{4}
\providecommand\BIBentryALTinterwordspacing{\spaceskip=\fontdimen2\font plus
\BIBentryALTinterwordstretchfactor\fontdimen3\font minus \fontdimen4\font\relax}
\providecommand\BIBforeignlanguage[2]{{%
\expandafter\ifx\csname l@#1\endcsname\relax
\typeout{** WARNING: IEEEtran.bst: No hyphenation pattern has been}%
\typeout{** loaded for the language `#1'. Using the pattern for}%
\typeout{** the default language instead.}%
\else
\language=\csname l@#1\endcsname
\fi
#2}}

\bibitem{mildenhall2020nerf}
B.~Mildenhall, P.~P. Srinivasan, M.~Tancik, J.~T. Barron, R.~Ramamoorthi, and R.~Ng, ``Nerf: Representing scenes as neural radiance fields for view synthesis,'' in \emph{European Conference on Computer Vision}.\hskip 1em plus 0.5em minus 0.4em\relax Springer, 2020, pp. 405--421.

\bibitem{Gao2022NeRFNR}
K.~Gao, Y.~Gao, H.~He, D.~Lu, L.~Xu, and J.~Li, ``Nerf: Neural radiance field in 3d vision, a comprehensive review,'' \emph{ArXiv}, vol. abs/2210.00379, 2022.

\bibitem{kerrevo}
J.~Kerr, L.~Fu, H.~Huang, Y.~Avigal, M.~Tancik, J.~Ichnowski, A.~Kanazawa, and K.~Goldberg, ``Evo-nerf: Evolving nerf for sequential robot grasping of transparent objects,'' in \emph{6th Annual Conference on Robot Learning}, 2022.

\bibitem{liu2023nerf}
J.~Liu, Q.~Nie, Y.~Liu, and C.~Wang, ``Nerf-loc: Visual localization with conditional neural radiance field,'' in \emph{{IEEE} International Conference on Robotics and Automation, {ICRA} 2023}.\hskip 1em plus 0.5em minus 0.4em\relax {IEEE}, 2023, pp. 9385--9392.

\bibitem{zhong23}
\BIBentryALTinterwordspacing
S.~Zhong, A.~Albini, O.~P. Jones, P.~Maiolino, and I.~Posner, ``Touching a nerf: Leveraging neural radiance fields for tactile sensory data generation,'' in \emph{Proceedings of The 6th Conference on Robot Learning}, ser. Proceedings of Machine Learning Research, K.~Liu, D.~Kulic, and J.~Ichnowski, Eds., vol. 205.\hskip 1em plus 0.5em minus 0.4em\relax PMLR, 14--18 Dec 2023, pp. 1618--1628. [Online]. Available: \url{https://proceedings.mlr.press/v205/zhong23a.html}
\BIBentrySTDinterwordspacing

\bibitem{abou2022implicit}
J.~Abou-Chakra, F.~Dayoub, and N.~S{\"u}nderhauf, ``Implicit object mapping with noisy data,'' \emph{arXiv preprint arXiv:2204.10516}, 2022.

\bibitem{muller2022instant}
T.~M{\"u}ller, A.~Evans, C.~Schied, and A.~Keller, ``Instant neural graphics primitives with a multiresolution hash encoding,'' \emph{ACM Transactions on Graphics (ToG)}, vol.~41, no.~4, pp. 1--15, 2022.

\bibitem{yang2021objectnerf}
B.~Yang, Y.~Zhang, Y.~Xu, Y.~Li, H.~Zhou, H.~Bao, G.~Zhang, and Z.~Cui, ``Learning object-compositional neural radiance field for editable scene rendering,'' in \emph{International Conference on Computer Vision ({ICCV})}, October 2021.

\bibitem{Locatello2020ObjectCentricLW}
F.~Locatello, D.~Weissenborn, T.~Unterthiner, A.~Mahendran, G.~Heigold, J.~Uszkoreit, A.~Dosovitskiy, and T.~Kipf, ``Object-centric learning with slot attention,'' \emph{ArXiv}, vol. abs/2006.15055, 2020.

\bibitem{Engelcke2021GENESISV2IU}
M.~Engelcke, O.~P. Jones, and I.~Posner, ``Genesis-v2: Inferring unordered object representations without iterative refinement,'' in \emph{Neural Information Processing Systems}, vol.~34, 2021, pp. 8085--8094.

\bibitem{lin2020space}
Z.~Lin, Y.~Wu, S.~V. Peri, W.~Sun, G.~Singh, F.~Deng, J.~Jiang, and S.~Ahn, ``{SPACE:} unsupervised object-oriented scene representation via spatial attention and decomposition,'' in \emph{8th International Conference on Learning Representations, {ICLR} 2020}.\hskip 1em plus 0.5em minus 0.4em\relax OpenReview.net, 2020.

\bibitem{Yu2021UnsupervisedDO}
H.~Yu, L.~J. Guibas, and J.~Wu, ``Unsupervised discovery of object radiance fields,'' in \emph{The Tenth International Conference on Learning Representations, {ICLR} 2022}.\hskip 1em plus 0.5em minus 0.4em\relax OpenReview.net, 2022.

\bibitem{Stelzner2021Decomposing3S}
K.~Stelzner, K.~Kersting, and A.~R. Kosiorek, ``Decomposing 3d scenes into objects via unsupervised volume segmentation,'' \emph{ArXiv}, vol. abs/2104.01148, 2021.

\bibitem{Wu2022ObPoseLC}
Y.~Wu, O.~P. Jones, and I.~Posner, ``Obpose: Leveraging canonical pose for object-centric scene inference in 3d,'' \emph{ArXiv}, vol. abs/2206.03591, 2022.

\bibitem{schwarz2020graf}
K.~Schwarz, Y.~Liao, M.~Niemeyer, and A.~Geiger, ``{GRAF}: Generative radiance fields for {3D}-aware image synthesis,'' \emph{Advances in Neural Information Processing Systems}, vol.~33, pp. 20\,154--20\,166, 2020.

\bibitem{niemeyer2021giraffe}
M.~Niemeyer and A.~Geiger, ``Giraffe: Representing scenes as compositional generative neural feature fields,'' in \emph{Proceedings of the IEEE/CVF Conference on Computer Vision and Pattern Recognition}, 2021, pp. 11\,453--11\,464.

\bibitem{zhou2023nerf}
A.~Zhou, M.~J. Kim, L.~Wang, P.~Florence, and C.~Finn, ``Nerf in the palm of your hand: Corrective augmentation for robotics via novel-view synthesis,'' \emph{arXiv preprint arXiv:2301.08556}, 2023.

\bibitem{sucar2021imap}
E.~Sucar, S.~Liu, J.~Ortiz, and A.~J. Davison, ``imap: Implicit mapping and positioning in real-time,'' in \emph{Proceedings of the IEEE/CVF International Conference on Computer Vision}, 2021, pp. 6229--6238.

\bibitem{ichnowski2021dex}
J.~Ichnowski, Y.~Avigal, J.~Kerr, and K.~Goldberg, ``Dex-nerf: Using a neural radiance field to grasp transparent objects,'' in \emph{Conference on Robot Learning, 2021}, ser. Proceedings of Machine Learning Research, vol. 164.\hskip 1em plus 0.5em minus 0.4em\relax {PMLR}, 2021, pp. 526--536.

\bibitem{dai2022graspnerf}
Q.~Dai, Y.~Zhu, Y.~Geng, C.~Ruan, J.~Zhang, and H.~Wang, ``Graspnerf: Multiview-based 6-dof grasp detection for transparent and specular objects using generalizable nerf,'' \emph{arXiv preprint arXiv:2210.06575}, 2022.

\bibitem{blukis2022neural}
V.~Blukis, T.~Lee, J.~Tremblay, B.~Wen, I.~S. Kweon, K.-J. Yoon, D.~Fox, and S.~Birchfield, ``Neural fields for robotic object manipulation from a single image,'' \emph{arXiv preprint arXiv:2210.12126}, 2022.

\bibitem{Burgess2019MONetUS}
C.~P. Burgess, L.~Matthey, N.~Watters, R.~Kabra, I.~Higgins, M.~M. Botvinick, and A.~Lerchner, ``Monet: Unsupervised scene decomposition and representation,'' \emph{ArXiv}, vol. abs/1901.11390, 2019.

\bibitem{Engelcke2019GENESISGS}
M.~Engelcke, A.~R. Kosiorek, O.~P. Jones, and I.~Posner, ``{GENESIS:} generative scene inference and sampling with object-centric latent representations,'' in \emph{8th International Conference on Learning Representations, {ICLR} 2020}.\hskip 1em plus 0.5em minus 0.4em\relax OpenReview.net, 2020.

\bibitem{wu2021apex}
Y.~Wu, O.~P. Jones, M.~Engelcke, and I.~Posner, ``Apex: Unsupervised, object-centric scene segmentation and tracking for robot manipulation,'' in \emph{2021 IEEE/RSJ International Conference on Intelligent Robots and Systems (IROS)}.\hskip 1em plus 0.5em minus 0.4em\relax IEEE, 2021, pp. 3375--3382.

\bibitem{yamada2023efficient}
J.~Yamada, J.~Collins, and I.~Posner, ``Efficient skill acquisition for complex manipulation tasks in obstructed environments,'' \emph{arXiv preprint arXiv:2303.03365}, 2023.

\bibitem{calli2015ycb}
B.~Calli, A.~Singh, A.~Walsman, S.~Srinivasa, P.~Abbeel, and A.~M. Dollar, ``The {YCB} object and model set: Towards common benchmarks for manipulation research,'' in \emph{2015 International Conference on Advanced Robotics (ICAR)}.\hskip 1em plus 0.5em minus 0.4em\relax IEEE, 2015, pp. 510--517.

\bibitem{johnson2017clevr}
J.~Johnson, B.~Hariharan, L.~Van Der~Maaten, L.~Fei-Fei, C.~Lawrence~Zitnick, and R.~Girshick, ``{CLEVR}: A diagnostic dataset for compositional language and elementary visual reasoning,'' in \emph{Proceedings of the IEEE conference on computer vision and pattern recognition}, 2017, pp. 2901--2910.

\bibitem{ester1996density}
M.~Ester, H.-P. Kriegel, J.~Sander, X.~Xu, \emph{et~al.}, ``A density-based algorithm for discovering clusters in large spatial databases with noise.'' in \emph{kdd}, vol.~96, no.~34, 1996, pp. 226--231.

\bibitem{ronneberger2015u}
O.~Ronneberger, P.~Fischer, and T.~Brox, ``U-net: Convolutional networks for biomedical image segmentation,'' in \emph{International Conference on Medical image computing and computer-assisted intervention}.\hskip 1em plus 0.5em minus 0.4em\relax Springer, 2015, pp. 234--241.

\bibitem{thomas2019kpconv}
H.~Thomas, C.~R. Qi, J.-E. Deschaud, B.~Marcotegui, F.~Goulette, and L.~J. Guibas, ``Kpconv: Flexible and deformable convolution for point clouds,'' in \emph{Proceedings of the IEEE/CVF international conference on computer vision}, 2019, pp. 6411--6420.

\bibitem{chan2022efficient}
E.~R. Chan, C.~Z. Lin, M.~A. Chan, K.~Nagano, B.~Pan, S.~De~Mello, O.~Gallo, L.~J. Guibas, J.~Tremblay, S.~Khamis, \emph{et~al.}, ``Efficient geometry-aware 3d generative adversarial networks,'' in \emph{Proceedings of the IEEE/CVF Conference on Computer Vision and Pattern Recognition}, 2022, pp. 16\,123--16\,133.

\bibitem{1981Random}
M.~A. Fischler and R.~C. Bolles, ``Random sample consensus: a paradigm for model fitting with applications to image analysis and automated cartography,'' \emph{Communications of the ACM}, 1981.

\bibitem{yu2021plenoctrees}
A.~Yu, R.~Li, M.~Tancik, H.~Li, R.~Ng, and A.~Kanazawa, ``Plenoctrees for real-time rendering of neural radiance fields,'' in \emph{Proceedings of the IEEE/CVF International Conference on Computer Vision}, 2021, pp. 5752--5761.

\bibitem{nerfstudio}
M.~Tancik, E.~Weber, E.~Ng, R.~Li, B.~Yi, J.~Kerr, T.~Wang, A.~Kristoffersen, J.~Austin, K.~Salahi, A.~Ahuja, D.~McAllister, and A.~Kanazawa, ``Nerfstudio: A modular framework for neural radiance field development,'' in \emph{ACM SIGGRAPH 2023 Conference Proceedings}, ser. SIGGRAPH '23, 2023.

\bibitem{Goodwin2021SemanticallyGO}
W.~Goodwin, S.~Vaze, I.~Havoutis, and I.~Posner, ``Semantically grounded object matching for robust robotic scene rearrangement,'' \emph{2022 International Conference on Robotics and Automation (ICRA)}, pp. 11\,138--11\,144, 2021.

\bibitem{Radford2021LearningTV}
A.~Radford, J.~W. Kim, C.~Hallacy, A.~Ramesh, G.~Goh, S.~Agarwal, G.~Sastry, A.~Askell, P.~Mishkin, J.~Clark, G.~Krueger, and I.~Sutskever, ``Learning transferable visual models from natural language supervision,'' in \emph{International Conference on Machine Learning}, 2021.

\bibitem{xiang2021learning}
Y.~Xiang, C.~Xie, A.~Mousavian, and D.~Fox, ``Learning rgb-d feature embeddings for unseen object instance segmentation,'' in \emph{Conference on Robot Learning}.\hskip 1em plus 0.5em minus 0.4em\relax PMLR, 2021, pp. 461--470.

\bibitem{Morrison2018ClosingTL}
D.~Morrison, J.~Leitner, and P.~Corke, ``Closing the loop for robotic grasping: {A} real-time, generative grasp synthesis approach,'' in \emph{Robotics: Science and Systems XIV, 2018}, 2018.

\bibitem{tosi2020distilled}
F.~Tosi, F.~Aleotti, P.~Z. Ramirez, M.~Poggi, S.~Salti, L.~D. Stefano, and S.~Mattoccia, ``Distilled semantics for comprehensive scene understanding from videos,'' in \emph{Proceedings of the IEEE/CVF conference on computer vision and pattern recognition}, 2020, pp. 4654--4665.

\bibitem{you2020keypointnet}
Y.~You, Y.~Lou, C.~Li, Z.~Cheng, L.~Li, L.~Ma, C.~Lu, and W.~Wang, ``Keypointnet: A large-scale 3d keypoint dataset aggregated from numerous human annotations,'' in \emph{Proceedings of the IEEE/CVF Conference on Computer Vision and Pattern Recognition}, 2020, pp. 13\,647--13\,656.

\bibitem{tatarchenko2019single}
M.~Tatarchenko, S.~R. Richter, R.~Ranftl, Z.~Li, V.~Koltun, and T.~Brox, ``What do single-view 3d reconstruction networks learn?'' in \emph{Proceedings of the IEEE/CVF conference on computer vision and pattern recognition}, 2019, pp. 3405--3414.

\end{thebibliography}
\end{document}